%% file: Arxiv_version.tex
\documentclass{article}
\usepackage{microtype}
\usepackage{graphicx}
\usepackage{booktabs}
\usepackage{multirow}
\usepackage[hypertexnames=false]{hyperref}

\usepackage[preprint]{preprint-layout}
\hypersetup{pdfsubject={Preprint}}
\makeatletter\renewcommand{\ICML@preprint}{\textit{Preprint.}}\makeatother
\usepackage{amsmath,amssymb,amsthm}
\usepackage{mathtools}
\usepackage[capitalize,noabbrev]{cleveref}

\input{math_commands.tex}

\usepackage{float}

\usepackage{colortbl}
\usepackage[most]{tcolorbox}
\newtcolorbox{promptbox}[1]{enhanced, breakable, colback=gray!4, colframe=black!55, boxrule=0.5pt, arc=1.5pt,
  left=5pt, right=5pt, top=4pt, bottom=4pt, fonttitle=\small\bfseries, coltitle=black, colbacktitle=gray!18,
  title=#1, before upper={\small\ttfamily\raggedright}, parbox=false}

\newtheorem{proposition}{Proposition}

\newcommand{\method}{\textsc{LBR}}

\icmltitlerunning{Recursively Self-Improving LLM Agents with Causality-Driven Exploration}

\begin{document}

\twocolumn[
  \icmltitle{Recursively Self-Improving LLM Agents with\\ Causality-Driven Exploration}
  \icmlsetsymbol{equal}{*}
  \icmlsetsymbol{intern}{$\dagger$}
  \begin{icmlauthorlist}
    \icmlauthor{Langzhou He}{equal,intern,uic}
    \icmlauthor{Qitian Wu}{equal,broad}
    \icmlauthor{Junhua Liu}{usc}
    \icmlauthor{Linjie Xu}{xlab}
    \icmlauthor{Philip S. Yu}{uic}
    \icmlauthor{David Wipf}{xlab}
  \end{icmlauthorlist}
  \icmlaffiliation{uic}{University of Illinois Chicago}
  \icmlaffiliation{broad}{Broad Institute of MIT and Harvard}
  \icmlaffiliation{usc}{University of Southern California}
  \icmlaffiliation{xlab}{Shanghai X-Lab, University of Hong Kong}
  \icmlcorrespondingauthor{David Wipf}{davidwipf@gmail.com}
  \icmlkeywords{LLM agents, causal inference, randomized experiments, relational databases, adaptive data analysis}
  \vskip 0.3in
]
\printAffiliationsAndNotice{\icmlEqualContribution $^\dagger$Work done during an internship at Shanghai X-Lab. }

\begin{abstract}
Recursively self-improving LLM agents propose candidates, verify them against a reward, and build on what works. Existing agents assume this reward can be queried repeatedly without consequence. In this paper, we show that this assumption fails in randomized experiments, where the outcomes that verify an agent's proposals are also the evidence it reports: reusing them invalidates the reported confidence intervals, and false positives reach 11\% where 5\% is expected. To address this, we propose \emph{Lock Before Reveal} (\method), a simple rule that fixes the analysis of each batch before its outcomes are revealed. The treatment-effect estimate then provably stays unbiased under any search policy, and in our tests false positives return to the level of an analysis with no search. Building on \method{}, we develop a self-improving agent that searches for programs that sharpen the estimate, with exploration driven by a causal reward. Across ten experiments built from real-world databases and twelve LLMs, it preserves valid coverage while using 100\% of the observations, reducing confidence interval widths by up to 44\% compared to unadjusted estimation. 

\end{abstract}

\section{Introduction}

Self-improving LLM agents follow a simple loop: propose a candidate, verify it, keep what works, and build the next proposal on what was kept \citep{romera2024,alphaevolve,dgm,huang2026memory}. The same loop now drives LLM agents that work as iterative data analysts: they inspect data sources \citep{huang2026deepresearch}, propose executable programs, evaluate them on observed labels, and revise later analyses from execution feedback \citep{relagent,shinn2023,madaan2023,yao2023,miao2026recodeh}. This loop is natural for prediction, where a reusable validation set can guide search and a separate test set evaluates the final predictor. However, when the same agents analyze randomized experiments to answer causal questions, the outcomes they need as feedback stop being freely reusable. When an agent tries multiple analyses on the same experimental outcomes and selects one based on its estimated effect, those outcomes serve twice: first to steer the search, and then as the statistical evidence reported. Randomization protects treatment assignment, but it does not correct for choosing an analysis after seeing its results \citep{simmons2011,gelman2014}, so the reported confidence intervals overstate the evidence for a treatment effect.

\begin{figure*}[t]
\centering
\includegraphics[width=\textwidth]{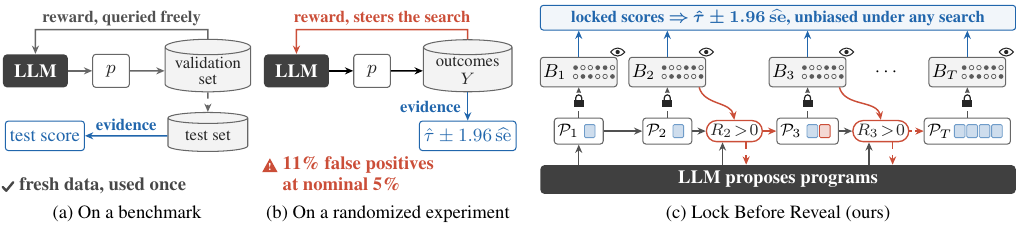}
\caption{Self-improvement when the reward is also the evidence. \textbf{(a)} On a benchmark, a validation set rewards the search, and a test set gives the evidence once. \textbf{(b)} In a randomized experiment, the same outcomes reward the search and serve as its evidence, which inflates false positives. \textbf{(c)} \method{} analyzes each batch with a pool locked before the batch is revealed, then uses its outcomes to reward the candidates of that round, and admits those with a positive reward into the next pool, so the estimate stays unbiased under any search.}
\label{fig:teaser}
\end{figure*}

In a randomized experiment, the agent searches for a better covariate adjustment. Adjusting for pre-treatment covariates that predict the outcome reduces estimator variance without biasing the treatment effect. These covariates are computed by \emph{programs} over the records of each unit before treatment, and good programs have to be searched for. Relational databases are a natural testbed: a customer, clinical study, post, or user is linked to historical events distributed across several tables \citep{relbench,relbenchv2,fey2024}, and relational predictors \citep{schlichtkrull2018,relgt} and foundation models \citep{tabpfn,kumorfm,kumorfm2,rdblearn} show that this linked history carries substantial predictive signal. A program is then a sequence of joins, temporal filters, and aggregations over pre-treatment history (e.g., the ratio of recent to long-run transaction value). LLM agents are well suited to search this space by reasoning over table and field descriptions, yet every round of outcome-guided refinement can invalidate the final confidence interval.

A familiar clinical example illustrates this failure mode: after a migraine drug trial yields a null overall effect, an analyst iteratively cuts the data by age and sex until reaching a subgroup with a significant effect. Reporting only this subgroup's ordinary 95\% confidence interval ignores the search that found it, so the interval no longer covers the true effect 95\% of the time. We observe the same failure when an LLM iteratively modifies its adjustment program on real data. On the RelBench Amazon dataset with randomly assigned treatment and zero true effect, we gave Qwen3.5-9B ten rounds to refine review-history features after observing the estimated effect and confidence interval. In individual runs, adaptive feature edits turned an initially null result into a false-positive interval excluding zero; across experiments, this search roughly doubled the false-positive rate relative to the LLM's first proposal, made before it saw any outcome.

\begin{figure*}[t]
\centering
\includegraphics[width=\textwidth,trim=8 146 35 151,clip]{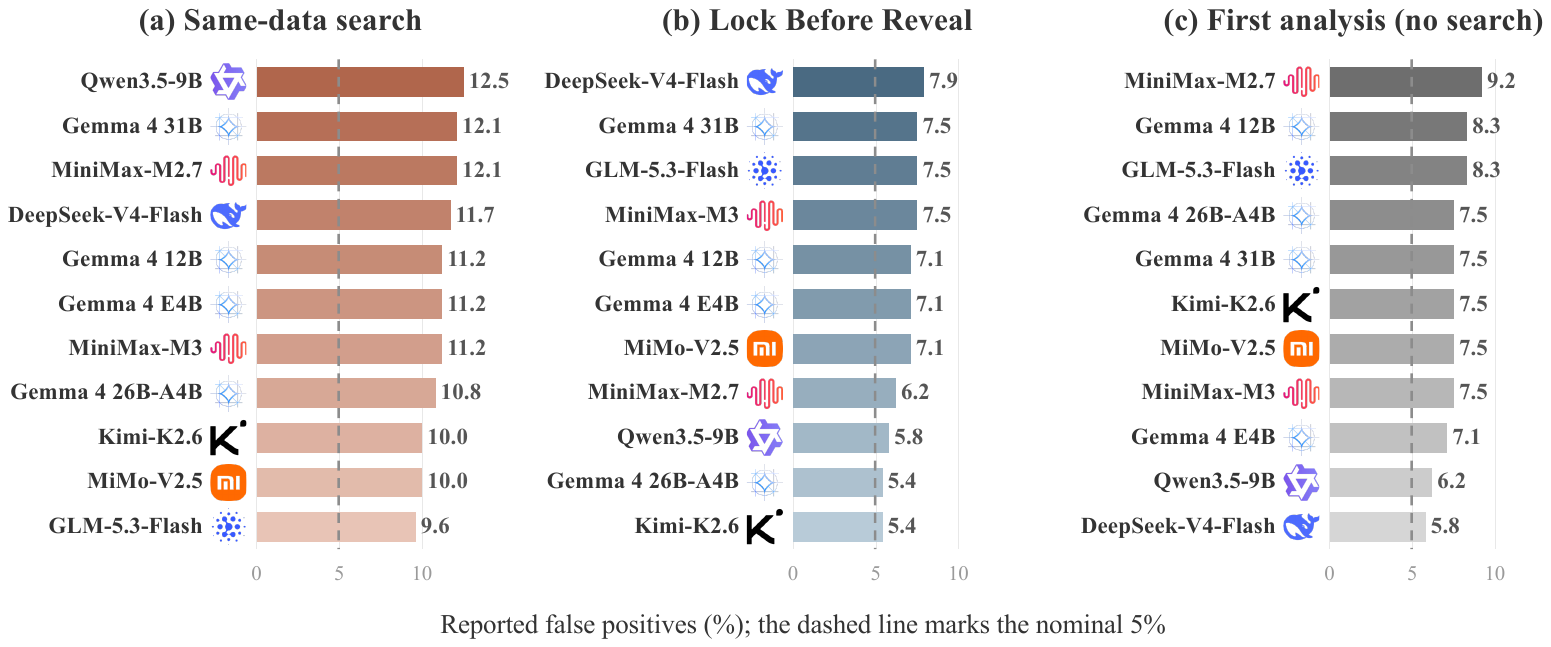}
\caption{Repeated search on the same outcomes produces false positives. Treatment was assigned at random with no true effect, and each LLM had ten rounds to analyze each experiment, changing its adjustment program after seeing the estimated effect and confidence interval. Bars show how often the reported 95\% interval excludes zero when a product team asks the LLM whether the effect is significant (four tasks, 240 experiments per LLM), and the dashed line marks the nominal 5\%. \textbf{(a)} Same-data search: every analysis uses the full experiment and the LLM chooses which one to report. \textbf{(b)} \method{}: each batch is analyzed by the program fixed before the batch's outcomes are revealed. \textbf{(c)} The LLM's first analysis, chosen before seeing any results. In the Gemma~4 names, E and A denote effective and active parameters. Pooled over the eleven LLMs, the three rates are 11.1\% (95\% interval 10.0--12.4\%), 6.8\% (5.9--7.8\%) and 7.5\%.}
\label{fig:fc1}
\end{figure*}

This raises a question for self-improving agents: can an agent improve itself on the very outcomes it will be judged by? To answer it, we introduce \emph{Lock Before Reveal} (\method). Figure~\ref{fig:teaser} contrasts it with self-improvement on a benchmark. The method is built on one timing rule: the analysis of a batch must be fixed before the batch's outcomes are revealed, and those outcomes may update only the analyses used on later batches. Under this rule, the same outcomes can safely do both jobs: they guide the search toward better programs, and every one of them still counts as valid causal evidence.

We instantiate \method{} in a recursively self-improving LLM agent. The agent's analysis is a pool: every program it has verified stays in the pool, and one outcome model fitted on all pooled feature columns is the locked analysis of the next batch. To guide its exploration, we introduce a marginal causal reward that measures how much a new candidate lowers this pooled model's prediction error on each newly revealed batch, the quantity that sets the variance of the treatment-effect estimator. Candidates with a positive reward join the pool, so each round's output is the next round's baseline, the bar rises as the agent improves, and the reward tells the LLM which information the pool still lacks. Our contributions are summarized as follows:

\begin{itemize}
    \item We formulate \emph{outcome-guided relational adjustment} as recursive self-improvement: an LLM searches for executable pre-treatment relational programs, every verified program joins a pooled adjustment model, and observed outcomes reward each proposal by what it adds to this model on fresh units. The pool's prediction error on units the agent never sees keeps falling.

    \item We introduce \emph{Lock Before Reveal}, a framework that allows outcome-guided search while preserving valid causal inference. Under known randomization probabilities, it keeps the treatment-effect estimate unbiased even as the LLM learns from earlier outcomes, and its confidence intervals remain valid in large samples.

    \item We provide a large-scale and comprehensive evaluation spanning ten tasks built from real relational databases and twelve LLM agents from eight model families. For every agent, \method{} preserves valid inference on every task, with intervals 13--17\% narrower than unadjusted estimation on average and 17--19\% narrower than separate-data search. In zero-effect experiments, its false positives stay at the level of an analysis with no search.
\end{itemize}

\section{Related Work}

\textbf{Self-improving agents with verifiable rewards.} Self-improving agents propose, verify, and keep what works\citep{schmidhuber2007}. With a verifier that scores every candidate, such agents have produced new programs and algorithms, as in FunSearch \citep{romera2024} and AlphaEvolve \citep{alphaevolve}, and have rewritten their own code, as in the Darwin G\"odel Machine \citep{dgm} and related systems \citep{stop,adas,godelagent}. Others improve later attempts from the feedback on earlier ones \citep{shinn2023,madaan2023} or accumulate verified skills in a growing library \citep{voyager}, and \citet{huang2026memory} and \citet{zou2026hac} review such self-evolving agents and the systems in which they work with people. In reinforcement-learning post-training, auxiliary signals likewise steer what an agent explores or trains on \citep{he2026actfocus,gu2026actorcurator}. These systems assume that the verifier's data can be queried at will. In a randomized experiment this does not hold, because the verifying data are also the evidence to be reported. Our agent uses each batch first as evidence, under an analysis fixed in advance, and only then as the reward for new candidates.

\textbf{Valid inference under adaptive data reuse.} Choosing an analysis after inspecting its results can make ordinary confidence intervals unreliable. Existing remedies account for how the analysis was selected \citep{lee2016,fithian2014}, control the information revealed through repeated queries \citep{dwork2015}, or separate the data used for selection from the data used for estimation \citep{athey2016,chernozhukov2018}. Related studies examine covariate selection in randomized experiments \citep{wager2016,lu2025}. In sequential settings, models fitted on past observations are evaluated on new ones \citep{dawid1984,luedtke2016}, while other approaches address adaptive treatment assignment \citep{hadad2021,zhang2020,kato2020,cook2024} and repeated statistical testing \citep{howard2021,ramdas2023}. Proposition~\ref{prop:valid} builds on this argument, but existing research pays less attention to LLM agents that continuously search for executable adjustment programs using outcome feedback.

\textbf{Automated data analysis.} A growing line of work automates feature engineering and prediction over tabular and relational data, through feature generation \citep{hollmann2023,kanter2015,wang2026corrective}, in-context foundation models \citep{tabpfn,kumorfm,kumorfm2,rdblearn}, and RelAgent \citep{relagent}, which extends LLM-driven program proposal and selection to relational schemas. These methods focus on predictive accuracy on a test set and do not address whether statistical inference stays valid once the outcomes under analysis also guide the search. For causal analysis, ORCA \citep{orca} relies on human-verified checkpoints, \citet{asher2026} show that the guardrails of LLM analysts against significance-seeking fail under simple prompt reframing, and \citet{thomas2026} explore preregistration for LLM-based analyses.

\section{Preliminaries}

\begin{figure*}[t]
\centering
\includegraphics[width=0.9\textwidth,trim=9 90 9 89,clip]{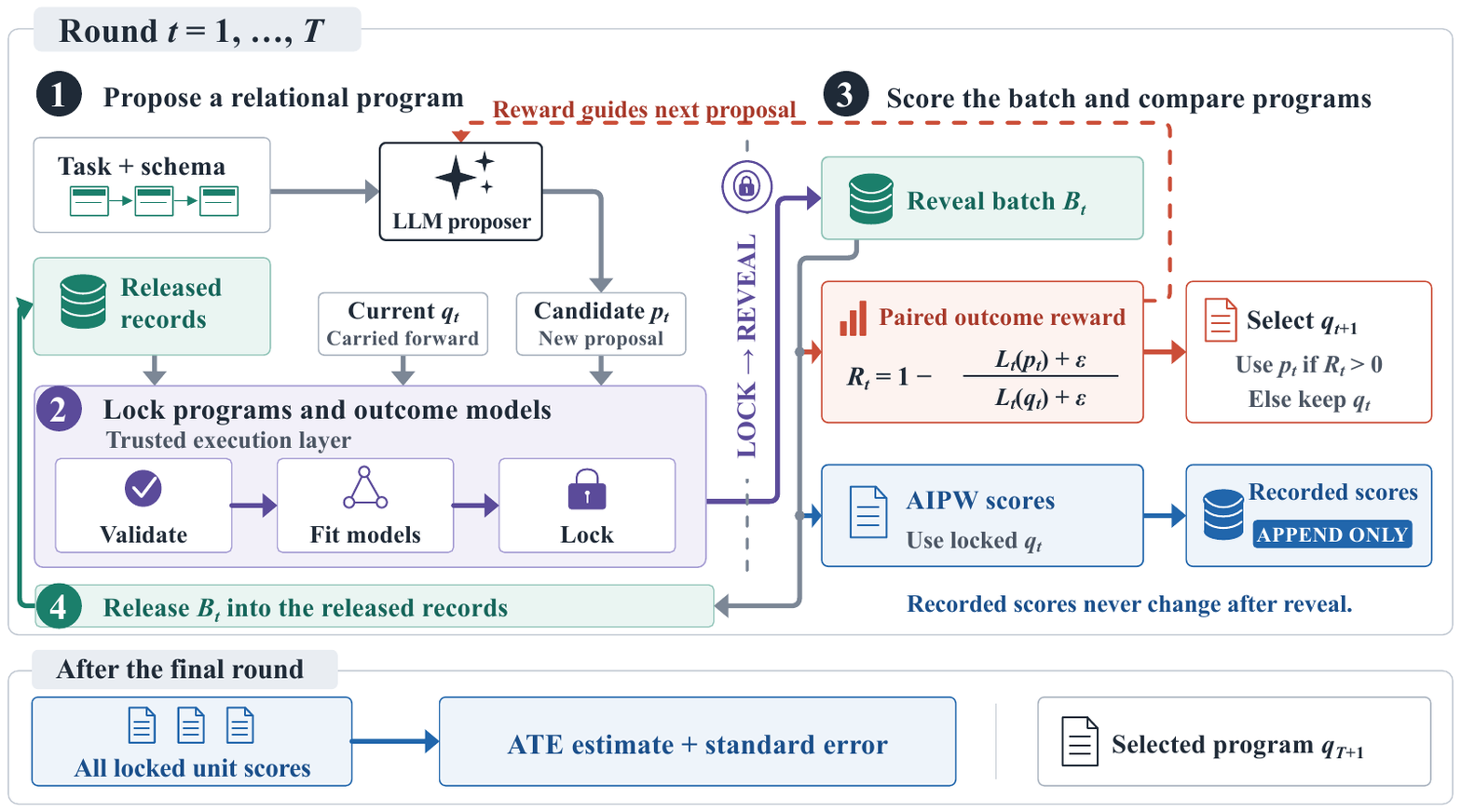}
\caption{One round of the agent. Before batch $B_t$ is revealed, the current pool $\mathcal P_t$ and the LLM's candidates are compiled, their outcome models are fitted on the released records, and everything is locked. After the reveal, the locked scores of $\mathcal P_t$ are permanently added to the ATE estimate, and the same outcomes give each candidate its marginal reward, which decides whether it joins $\mathcal P_{t+1}$ and is returned to the LLM for the next round. After the final round, all locked scores aggregate into the ATE estimate and standard error.}
\label{fig:pipeline}
\end{figure*}

\label{sec:setup}

\paragraph{Causal effects in randomized experiments.}
An experiment has units $i=1,\ldots,N$, such as users, and unit $i$ enters the experiment at a known time $c_i$. Each unit is assigned to treatment ($A_i=1$) or control ($A_i=0$) with a probability $e_i=\Pr(A_i=1)$ that the experimenter sets and therefore knows. Let $Y_i(1)$ and $Y_i(0)$ denote the potential outcomes of unit $i$ under treatment and control; the experiment reveals only $Y_i=A_iY_i(1)+(1-A_i)Y_i(0)$. The target is the average treatment effect (ATE) $\tau=\mathbb{E}[Y_i(1)-Y_i(0)]$. With equal assignment probabilities, the difference in mean outcomes between the two groups is already unbiased for $\tau$, and covariates serve only to reduce its variance.

\paragraph{Adjustment programs.}
An \emph{adjustment program} $p$ maps the data recorded about the units to a feature vector $X_i^{(p)}$ for every unit, for example the number of orders a user placed in the 90 days before entering the experiment. Every event a program uses must precede the unit's entry time $c_i$: later events may already be affected by the treatment, and adjusting for them would bias the effect estimate. In our experiments the data are a relational database $\mathcal{D}=\{\mathcal{T}_1,\ldots,\mathcal{T}_K\}$, a set of tables linked by primary--foreign key relations \citep{relbench,relagent}, and the units are the rows of one entity table. A program may then traverse foreign keys, restrict events to a time window, filter rows, aggregate values, and combine several aggregates.

\paragraph{Variance-reducing scores.}
We adjust through the augmented inverse probability weighting (AIPW) score \citep{robins1994,bang2005}, which uses covariates without risking bias. For a program $p$, let $\hat\mu_{1}^{(p)}$ and $\hat\mu_{0}^{(p)}$ be outcome models for treated and control units that take the program's features as input. The AIPW score of unit $i$ is
\begin{equation}
\begin{split}
\psi_{i}(p) ={}& \hat\mu_{1}^{(p)}(X_i^{(p)})\!-\!\hat\mu_{0}^{(p)}(X_i^{(p)})
\!+\!\frac{A_i\{Y_i-\hat\mu_{1}^{(p)}(X_i^{(p)})\}}{e_i}\\[2pt]
&-\frac{(1-A_i)\{Y_i-\hat\mu_{0}^{(p)}(X_i^{(p)})\}}{1-e_i}.
\end{split}
\label{eq:aipw}
\end{equation}
If the program and outcome models do not depend on the outcomes being analyzed, randomization makes the two correction terms cancel the models' errors on average, so the average score is unbiased for $\tau$ even when the models are misspecified; more accurate models lower its variance and shorten the confidence interval. In our setting the outcomes also guide the search over programs, and the question is when a program chosen this way can still support valid inference.

\section{The Lock Before Reveal Agent}
\label{sec:method}

The agent has three modules: the inference rule under which every batch enters the estimate (Section~\ref{sec:inference}), a pooled analysis with a marginal causal reward that verifies the proposed candidates and guides subsequent ones (Section~\ref{sec:reward}), and an LLM that proposes adjustment programs (Section~\ref{sec:proposer}). Figure~\ref{fig:pipeline} shows how they work together. The design principle is a timing rule: the analysis of any batch must be completely fixed before that batch's treatment assignments and outcomes are revealed. Feedback from the batch updates only the analysis used on future batches.

We divide the experimental units into $T$ sequential batches $B_1, \ldots, B_T$. Before round $t$, the agent holds the released records $\mathcal R_{t-1}$, which are the pre-experiment data and the batches revealed so far, and the pool $\mathcal P_t$ of programs admitted so far, starting from one program proposed from the database schema alone. Each round has four steps (Algorithm~\ref{alg:main}). \textbf{Propose:} from round 2 onward, the LLM proposes $k$ candidate programs $p_{t,1}, \ldots, p_{t,k}$. \textbf{Lock:} the system fits outcome models for $\mathcal P_t$ and for the pool with each candidate added, $\mathcal P_t \cup \{p_{t,j}\}$, on $\mathcal R_{t-1}$, and locks all of them. \textbf{Reveal:} batch $B_t$ is revealed and scored by the locked models of $\mathcal P_t$; these scores enter the treatment-effect estimate and are never revised. \textbf{Reward:} the same outcomes give each candidate a marginal reward, and candidates with a positive reward join the pool to form $\mathcal P_{t+1}$. After round $T$, the recorded scores are averaged into the average treatment effect (ATE) estimate and its standard error. The LLM stays fixed throughout; what changes from round to round is the pool and the feedback given to the LLM.

\begin{algorithm}[t]
\caption{The Lock Before Reveal agent}
\label{alg:main}
\begin{algorithmic}[1]
\STATE {\bfseries Input:} pre-experiment data $\mathcal R_0$; batches $B_1,\ldots,B_T$; initial pool $\mathcal P_1$ holding the LLM's first program, proposed from the schema
\FOR{$t=1,\ldots,T$}
    \STATE \textbf{Propose:} the LLM proposes candidates $p_{t,1},\ldots,p_{t,k}$ (none in round 1); invalid ones are discarded
    \STATE \textbf{Lock:} fit the outcome models $\hat\mu_{1,t}^{(\mathcal P)}$ and $\hat\mu_{0,t}^{(\mathcal P)}$ on $\mathcal R_{t-1}$ for $\mathcal P=\mathcal P_t$ and for each $\mathcal P=\mathcal P_t\cup\{p_{t,j}\}$; lock all programs and models
    \STATE \textbf{Reveal:} reveal $B_t$ and permanently record the scores $\psi_{i,t}(\mathcal P_t)$, $i\in B_t$
    \STATE \textbf{Reward:} compute $R_t(p_{t,j})$ by \eqref{eq:marginal-reward} and return the rewards and admission decisions to the LLM
    \STATE $\mathcal P_{t+1}\leftarrow\mathcal P_t\cup\{p_{t,j}:R_t(p_{t,j})>0\}$ and $\mathcal R_t\leftarrow\mathcal R_{t-1}\cup B_t$
\ENDFOR
\STATE {\bfseries Output:} $\hat\tau$ and $\widehat{\mathrm{SE}}$ from \eqref{eq:estimator}; the final pool $\mathcal P_{T+1}$
\end{algorithmic}
\end{algorithm}

\subsection{Valid inference from locked analyses}
\label{sec:inference}

\begin{figure}[t]
\centering
\includegraphics[width=\columnwidth]{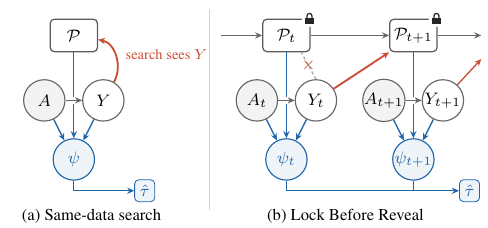}
\caption{Causal graphs of two ways to reuse outcomes. \textbf{(a)} Same-data search chooses $\mathcal P$ after seeing the outcomes $Y$. \textbf{(b)} \method{} locks $\mathcal P_t$ before $A_t$ and $Y_t$ are revealed, so $Y_t$ reaches only $\mathcal P_{t+1}$ and every $\psi_t$ stays centered at its batch's treatment effect.}
\label{fig:causal}
\end{figure}

The analysis of a batch is not one program but the pool of every program admitted so far. To turn a set of programs $\mathcal P$ into one model, we collect the feature columns of all its programs, with duplicate columns dropped, into a feature vector $X_i^{(\mathcal P)}$ for each unit $i$. We then fit two outcome models on the released records $\mathcal R_{t-1}$, both ridge regressions of the outcome on these features: $\hat\mu_{1,t}^{(\mathcal P)}$ on treated units and $\hat\mu_{0,t}^{(\mathcal P)}$ on control units.
The pre-experiment data $\mathcal R_0$ are assumed to contain randomized units, as in a pilot phase; in our experiments they are the training split (Appendix~\ref{app:setup}).
The score $\psi_{i,t}(\mathcal P_t)$ of each unit $i\in B_t$ is computed as in \eqref{eq:aipw} with the features and models of $\mathcal P_t$, so the analysis of a batch depends only on data available before its reveal.

\begin{table*}[t]
\caption{Interval width relative to no adjustment for twelve LLMs, averaged over ten RelBench tasks with 30 seeds each. Lower is better. Without an LLM, CUPED reaches 0.874 on the eight tasks with pre-period outcomes and DFS reaches 0.945. The first program is proposed from the schema alone, separate-data search selects programs on batches 1--3 and estimates from batches 4--10, and the one-program variant keeps a single program in place of the pool. Coverage is the share of nominal 95\% intervals that contain $\tau$, gain is the width reduction over separate-data search, and every \method{} agent passes the coverage check on every task.}
\label{tab:baselines}
\begin{center}
\small
\setlength{\tabcolsep}{2.65pt}
\input{tables/mm_main.tex}

\end{center}
\end{table*}

\begin{figure*}[t]
\begin{center}
\includegraphics[width=\textwidth,trim=21 161 12 145,clip]{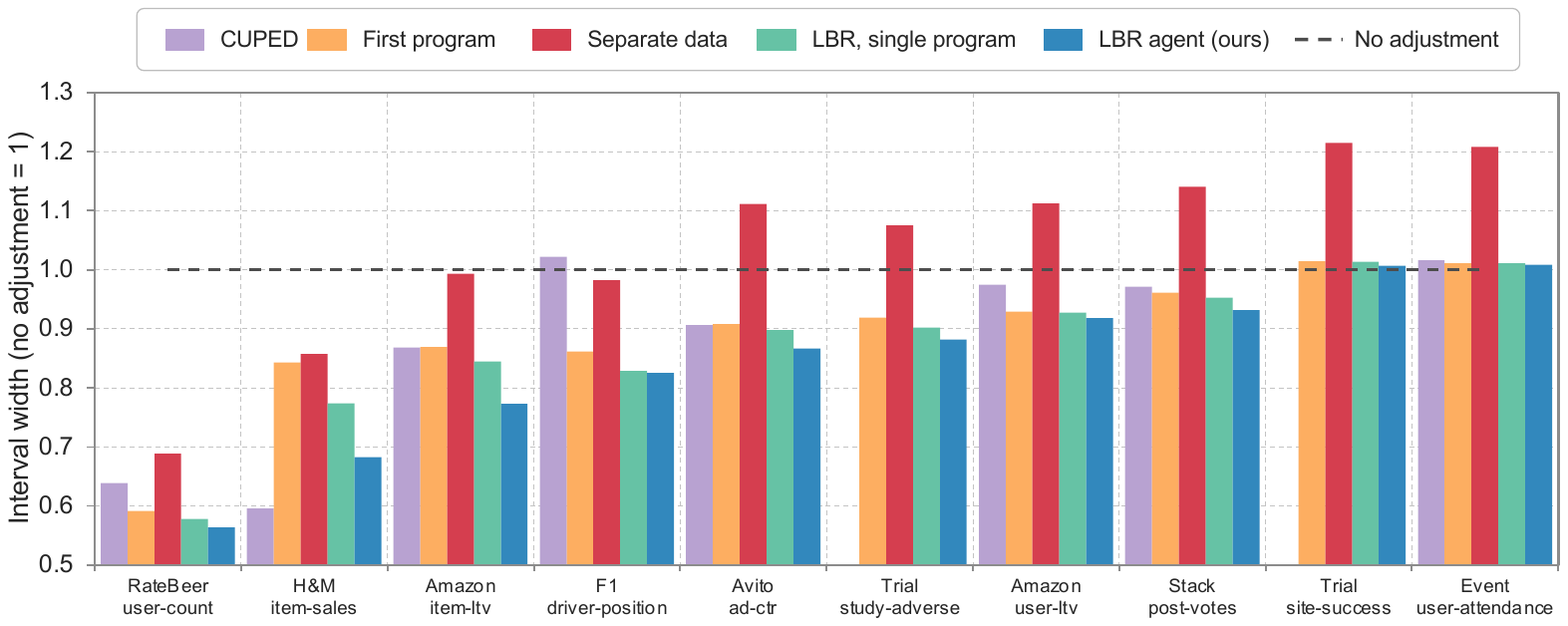}
\end{center}
\caption{Interval width by task with GPT-5.6 Luna, relative to no adjustment (dashed line; 30 seeds per task). Lower is better. Single program: \method{} with one program in place of the pool (Table~\ref{tab:ablation}). Tasks are ordered by the agent's width, and CUPED is missing on the two tasks without a pre-period outcome.}
\label{fig:tasks}
\end{figure*}

The outcomes of a batch are used twice: they reward the candidates, and they are the evidence for the treatment effect. The two uses must not interact within a batch, because if the outcomes of $B_t$ could influence how $B_t$ itself is analyzed, the analysis would be selected after seeing its results. We therefore fix $\mathcal P_t$ and its outcome models before the treatment assignments and outcomes in $B_t$ are revealed. Formally, let $\mathcal F_{t-1}$ denote everything known before this reveal, including earlier data, rewards, and the locked pools and models (Appendix~\ref{app:proof}); the timing rule requires $\mathcal P_t$, $\hat\mu_{1,t}^{(\mathcal P_t)}$, and $\hat\mu_{0,t}^{(\mathcal P_t)}$ to depend only on $\mathcal F_{t-1}$. The candidates and the models that include them are locked at the same point, so every program is evaluated on outcomes it has not seen, and any admission based on these outcomes takes effect only in the next batch. Figure~\ref{fig:causal} shows this ordering as a causal graph.

Every batch is scored once by its locked pool and models, so no batch needs to be reserved for search. Recall that $\psi_{i,t}(\mathcal P_t)$ is the locked score of unit $i\in B_t$, let $N=\sum_{t=1}^T n_t$, and let $s_t^2$ be the sample variance of the locked scores within batch $t$. The final estimate averages all recorded scores, and its estimated standard error is given by
\begin{equation}
\hat\tau=\frac{1}{N}\sum_{t=1}^T\sum_{i\in B_t}\psi_{i,t}(\mathcal P_t),
\qquad
\widehat{\mathrm{SE}}^2=\frac{1}{N^2}\sum_{t=1}^T n_t s_t^2.
\label{eq:estimator}
\end{equation}

To state the guarantee, let $\tau_i=Y_i(1)-Y_i(0)$ be the treatment effect for unit $i$, and let $\bar\tau_{B_t}=n_t^{-1}\sum_{i\in B_t}\tau_i$ be the average effect in batch $B_t$. The target for the full experiment is $\tau_N=N^{-1}\sum_{t=1}^T n_t\bar\tau_{B_t}$. Let $\mathcal G_{t-1}$ contain $\mathcal F_{t-1}$ together with the potential outcomes of all experimental units.

\begin{proposition}
\label{prop:valid}
Assume that for every round $t$: (i) conditional on $\mathcal G_{t-1}$, treatment assignments in $B_t$ are independent with known probabilities bounded away from zero and one; (ii) $\mathcal P_t$, $\hat\mu_{1,t}^{(\mathcal P_t)}$, and $\hat\mu_{0,t}^{(\mathcal P_t)}$ depend only on $\mathcal F_{t-1}$; and (iii) the scores $\psi_{i,t}(\mathcal P_t)$ have finite variance. Then the average score of every batch is conditionally centered at that batch's treatment effect, $\mathbb{E}[\bar\psi_{B_t}(\mathcal P_t)\mid\mathcal{G}_{t-1}]=\bar\tau_{B_t}$, so $\hat\tau$ is unbiased for $\tau_N$ no matter how the search policy builds pools from the released history. Under the martingale regularity conditions stated in Appendix~\ref{app:proof}, $(\hat\tau-\tau_N)/\widehat{\mathrm{SE}}$ is asymptotically standard normal when treatment effects are constant, and $\widehat{\mathrm{SE}}$ in \eqref{eq:estimator} is asymptotically conservative otherwise, so $\hat\tau\pm z_{1-\alpha/2}\widehat{\mathrm{SE}}$ covers $\tau_N$ with asymptotic probability at least $1-\alpha$ as $N\to\infty$.
\end{proposition}

Intuitively, once the analysis of $B_t$ is locked, the only randomness left is the treatment assignment of the new batch, and for any fixed outcome model, randomization makes the correction terms of \eqref{eq:aipw} cancel the model's error on average. Proposition~\ref{prop:valid} therefore does not depend on how the LLM proposes programs or how the reward guides the search, and it holds for the single-program analysis of Appendix~\ref{app:single} and the pooled analysis alike. Better pools improve precision, but the unbiasedness of the estimate does not depend on the LLM finding them. For the same reason, the formulas in \eqref{eq:estimator} applied to the batches revealed so far give an estimate and interval at any round fixed in advance. The full proof and finite-sample checks are provided in Appendices~\ref{app:proof} and \ref{app:validity}.

\subsection{Causality-driven exploration}
\label{sec:reward}

Candidates are tested by what they add to the pooled model. Before the reveal, we also fit and lock the two outcome models for the pool with each candidate added, $\mathcal P_t\cup\{p_{t,j}\}$. After the reveal, we measure the prediction error of each set of programs $\mathcal P$ on the new batch. A unit's outcome is observed only under the treatment it actually received, so treated units are compared with the predictions of $\hat\mu_{1,t}^{(\mathcal P)}$ and control units with those of $\hat\mu_{0,t}^{(\mathcal P)}$. With $n_t=|B_t|$, the resulting \emph{factual loss} is
\begin{equation}
\begin{split}
L_t(\mathcal P)={}&\frac{1}{n_t}\sum_{i\in B_t}\Big[
A_i\{Y_i-\hat\mu_{1,t}^{(\mathcal P)}(X_i^{(\mathcal P)})\}^2\\
&+(1-A_i)\{Y_i-\hat\mu_{0,t}^{(\mathcal P)}(X_i^{(\mathcal P)})\}^2\Big].
\end{split}
\label{eq:factual-loss}
\end{equation}
The \emph{marginal reward} of candidate $p_{t,j}$ is the relative reduction of this loss when the candidate is added to the pool,
\begin{equation}
R_{t}(p_{t,j})=1-\frac{L_t(\mathcal P_t\cup\{p_{t,j}\})+\epsilon}{L_t(\mathcal P_t)+\epsilon},
\qquad \epsilon=10^{-6}.
\label{eq:marginal-reward}
\end{equation}
A positive reward means that the pool predicts the new outcomes better with the candidate than without it, and $|R_t(p_{t,j})|$ is the relative size of the difference. Every candidate of a round is compared with the same pool on the same units, so candidates are judged on equal terms, and the ratio removes the scale of the outcome. Candidates with $R_t(p_{t,j})>0$ are admitted, $\mathcal P_{t+1}=\mathcal P_t\cup\{p_{t,j}: R_t(p_{t,j})>0\}$, and the outcome models of the new pool are fitted on the enlarged released records in the next round.

The reward in \eqref{eq:marginal-reward} is designed to align the search with the goal of the experiment, a precise estimate of the treatment effect. In a randomized experiment, the more accurately the outcome models predict, the smaller the variance of the treatment-effect estimate, so a lower loss on new units generally means a narrower confidence interval. A candidate is therefore rewarded for sharpening the causal estimate, and the gain is measured on units that no model has seen. We call the reward \emph{causal} for this reason. The reward is also \emph{recursive}, because its baseline is the agent's own current pool: as the pool improves, a candidate earns a reward only for what the pool does not already contain. A program that would have been valuable in round 2 may earn nothing in round 8 if the pool already carries its information, which steers the agent toward information the pool still lacks. Finally, an admitted program is never removed and a rejected candidate leaves the analysis unchanged, so the gains of earlier rounds are preserved, and a predictor known in advance, such as a pre-period outcome, can be placed in $\mathcal P_1$ as a system-provided member.

For each candidate, the LLM receives the sign and size of its reward, whether it was admitted, and which of its features received the largest weight in the model. It does not see batch outcomes, ATE estimates, standard errors, or confidence intervals. Validity does not depend on this choice: under the timing rule of Section~\ref{sec:inference}, the estimate stays unbiased even when the LLM sees the estimate and interval after every round (Section~\ref{sec:fc1}).

\subsection{LLM-based program proposal}
\label{sec:proposer}

From round 2 onward, the LLM receives the task description, the database schema with table and field definitions, the current pool members with their rewards, and the previously rejected candidates. It returns a candidate program in a predefined typed language rather than free-form SQL. A program specifies which table links to follow, which time windows and filters to apply, and how to aggregate and combine the resulting values. We sample $k=4$ candidates per round from the same prompt and evaluate each one separately. The prompt asks for information the pool does not have yet, such as other relations, time windows, aggregations, or transforms, so each round's candidates build on what the pool already contains.

A validator checks that the program follows the language rules. A deterministic SQL generator then computes its features, restricting all events to those recorded before each unit's entry time $c_i$. The same validation and compilation steps apply to the first program and to every candidate. Candidates that break the language rules or whose queries exceed the computation budget are reported back to the LLM with the reason and are not evaluated. The full language and prompts are given in Appendix~\ref{app:prompts}.

\begin{figure*}[t]
\begin{center}
\includegraphics[width=\textwidth,trim=5 202 7 191,clip]{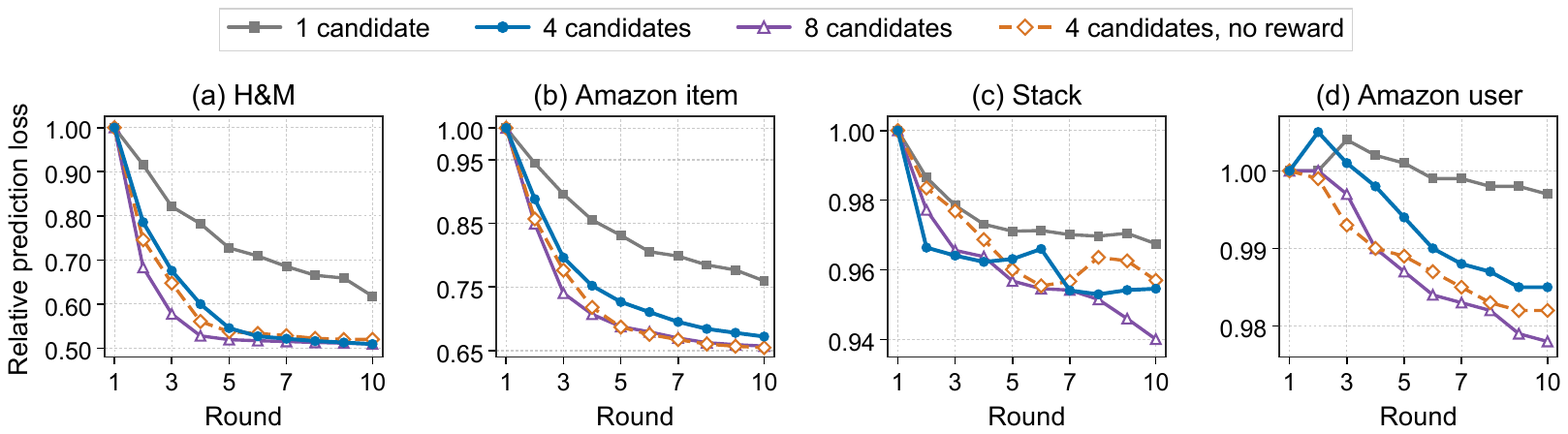}
\end{center}
\caption{The analysis improves while the experiment runs (GPT-5.6 Luna, 30 seeds per task). Prediction loss of the pool that will analyze the next batch, measured on a diagnostic batch the agent never sees and relative to the round-1 pool, by round, on the four tasks from the three largest databases: \textbf{(a)} H\&M item-sales, \textbf{(b)} Amazon item-ltv, \textbf{(c)} Stack post-votes, and \textbf{(d)} Amazon user-ltv, for one, four and eight candidates per round and for four candidates without the reward. Lower is better.}
\label{fig:selfimprove}
\end{figure*}

\section{Experiments}
\label{sec:exp}

We evaluate the agent on randomized experiments built from real relational databases, in which the true effect is known. Relational databases provide real outcomes and a large space of adjustment programs, so a good analysis has to be searched for. We first show that searching on the same outcomes inflates false positives and that \method{} removes the inflation, then compare the agent's precision with other ways of using the same data, examine how its analysis improves during the experiment, and repeat the study with twelve LLMs.

\subsection{Experimental setup}

\textbf{Data and known treatment effects.} We use ten regression tasks from RelBench and RelBench v2 \citep{relbench,relbenchv2}. The units, their histories, and their future outcomes come from real databases. Following designs used to evaluate covariate adjustment \citep{deng2013,guo2021}, we assign treatment at random and add a known effect. For a unit with observed future outcome $y_i$, let $Z_i=\{\log(1+y_i)-\mu_{\mathrm{train}}\}/\sigma_{\mathrm{train}}$ be its standardized log outcome, with mean and standard deviation computed on the training split. We set $Y_i(0)=Z_i$, draw $A_i\sim\mathrm{Bern}(1/2)$ independently of the unit histories, and observe $Y_i=Z_i+A_i$ \citep{imbens2015}, so the true ATE is $\tau=1$. Adjustment programs may use only information recorded before each unit's entry time.

\begin{table}[t]
\caption{Variants of the agent with GPT-5.6 Luna across ten tasks (30 seeds per task). Width is relative to unadjusted estimation, and Features denotes the mean number of feature columns in the final pool. \method{} uses four candidates per round by default.}
\label{tab:ablation}
\centering
\small
\setlength{\tabcolsep}{3pt}
\renewcommand{\arraystretch}{1.12}
\begin{tabular*}{\columnwidth}
{@{\extracolsep{\fill}}lcc@{}}
\toprule
Variant & Width $\downarrow$ & Features $\downarrow$ \\
\midrule
Single program ($k=1$)
    & 0.873 & -- \\
w/ 1 candidate
    & 0.862 & 36 \\
w/ 8 candidates
    & 0.840 & 153 \\
w/o reward
    & 0.842 & 179 \\
w/ CUPED initialization
    & 0.834 & 81 \\
\midrule
\textbf{LBR (Ours)}
    & 0.846 & 95 \\
\bottomrule
\end{tabular*}
\end{table}

\textbf{Batches and seeds.} The training split serves as pre-experiment data: it is used to fit the initial outcome models and does not enter the ATE estimate. The validation split is divided, without inspecting outcomes, into ten batches of up to 300 units, plus one diagnostic batch that is never shown to the agent. Each experiment is repeated with 30 seeds, and all methods share the same treatment assignments and batches.

\textbf{LLMs and agent settings.} We use twelve LLMs from eight model families: GPT-5.6 Luna \citep{openai2026gpt56}, Qwen3.5-9B \citep{qwen2026qwen35}, Gemma~4 (E4B, 12B, 26B-A4B, and 31B) \citep{gemmateam2026gemma4}, MiniMax M2.7 \citep{minimax2026m27} and M3 \citep{lai2026msa}, DeepSeek-V4-Flash \citep{deepseek2026v4}, GLM-5.3-Flash \citep{glm2026glm5}, MiMo-V2.5 \citep{xiaomi2026mimo25}, and Kimi-K2.6 \citep{moonshot2026kimik26}. Unless stated otherwise, every experiment is run with with the same prompts and the same agent settings. Each run has $T=10$ rounds with $k=4$ candidates per round from round 2 onward, and all configurations of a seed share the first program. The outcome models are the ridge regressions of Section~\ref{sec:inference}.

\textbf{Compared methods.} We compare the agent with five baselines. Three fix the analysis before any outcome is seen: CUPED \citep{deng2013}, DFS \citep{kanter2015}, and the first program proposed by the LLM. Two let outcomes guide the search: separate-data search \citep{cox1975} and \method{} with a single program.

\textbf{Metrics.} For validity we report how often the 95\% confidence interval contains the true effect and, in experiments with no effect, how often it excludes zero. As a coverage check, we require at least 26 of the 30 intervals of every task to contain the true effect. For precision we report the interval width divided by the width with no adjustment on the same task, so values below one are narrower, and we average these ratios with equal weight per task.

Appendix~\ref{app:setup} gives the task definitions, the construction of the batches, and the implementation of each baseline, and Appendix~\ref{app:pooled} the agent settings.

\subsection{Searching on the same outcomes inflates false positives}
\label{sec:fc1}

We first show the problem that the timing rule solves. We remove the treatment effect from the experiments ($Y_i=Z_i$), so that any significant result is a false positive. An LLM analyst then searches for an adjustment program over ten rounds, seeing the estimate and $p$-value of each program it tries, and is asked to tell a product team whether the effect is significant. The analyst either works on the complete data and reports its final program, or works under \method{}, where the program of round $t$ is fixed before batch $t$ is revealed.

We run this study with eleven LLMs, each on 240 experiments in both settings, as shown in Figure~\ref{fig:fc1}. Searching on the complete data doubles the false-positive rate: instead of the expected 5\%, the reported interval excludes zero in 9.6--12.5\% of experiments, depending on the LLM. Under \method{} the same LLMs are at 5.4--7.9\%, the level of their first program, which involves no search at all. The protection comes from the order of events: when the programs proposed under \method{} are re-analyzed on the complete data and the most significant one is reported, the rate returns to 8.8--12.1\%. With a true effect, the agent's intervals pass the coverage check for every LLM (Table~\ref{tab:baselines}), and a larger check with 2,000 fresh treatment assignments gives a coverage of 94.9--95.0\% (Appendix~\ref{app:validity}). Appendix~\ref{app:fc1} gives the details of the zero-effect study.

\subsection{Precision across relational tasks}
\label{sec:main}

After establishing that \method{} keeps the search valid, we want to know how much precision the search brings. We return to the experiments with the known effect $\tau=1$, and Table~\ref{tab:baselines} compares the agent with other ways of analyzing them, for each of the twelve LLMs. The agent gives the narrowest intervals for every LLM, 13--17\% narrower than without adjustment. The reduction scales with how well the relational history predicts the outcome and, with GPT-5.6 Luna, reaches 44\% on RateBeer and 32\% on H\&M (Figure~\ref{fig:tasks}).

Separate-data search, the standard safe way to search, is wider than no adjustment for every LLM, by 2--7\%, because the three batches it sets aside are worth more than the adjustment it finds. Under \method{} every batch both improves the analysis and enters the estimate, and the agent's intervals are 17--19\% narrower than those of separate-data search, on every task for every LLM. The agent is also 3--6\% narrower than the LLM's first program, so the programs added in later rounds carry information that the first program does not.

Programs discovered by the agent already outperform the covariate that experimenters specify by hand. Given only table and field descriptions, the GPT-5.6 Luna agent is narrower than CUPED on seven of the eight tasks that have a pre-period outcome. Combining the two helps further: warm-starting the pool with the pre-period outcome narrows the mean width from 0.846 to 0.834 (Table~\ref{tab:ablation}) and the H\&M interval from 0.683 to 0.593.

\subsection{The analysis improves while the experiment runs}
\label{sec:selfimprove}
\label{sec:feedback}

The precision above is the result of ten rounds of search, so we next examine whether the analysis indeed improves from round to round. After every round we take the pool that will analyze the next batch and measure how well it predicts the outcomes of units that the agent never sees. Its prediction error falls round after round: by round 10 it is 49\% lower than that of the first program alone on H\&M (Figure~\ref{fig:selfimprove}). This improvement does not cost any data, because each batch is first analyzed by a pool fixed beforehand and only then used to improve the next one, and the estimate is available throughout the experiment.

We then ask what the causal reward contributes to this search, and run the agent without it: every candidate is admitted and the LLM sees only its past proposals. This variant reaches a similar final width, but its pool is much larger (179 feature columns against 95) and it stops improving after round 6. The marginal causal reward reaches this precision with about half the feature columns (Table~\ref{tab:ablation}). The pool also matters: keeping one program at a time in place of the pool, as the analyst of Section~\ref{sec:fc1} does, gives 0.873.

\subsection{Robustness of \method{} across LLMs}
\label{sec:llms}

All results so far rely on an LLM to propose the programs, so we finally test whether they depend on which LLM is used. We repeat the zero-effect study with eleven LLMs and the precision experiments with twelve (Figure~\ref{fig:fc1} and Table~\ref{tab:baselines}). The conclusions hold for every LLM: the intervals pass the coverage check on every task, and the agent is narrower than separate-data search on every task, by 17--19\% on average. The choice of LLM changes only how much precision is found, from 0.830 to 0.874 of the unadjusted width.

\section{Conclusion}

We study self-improving LLM agents whose reward is also the evidence they report. In this setting, observations may improve future analyses, but they cannot choose the analysis used to evaluate themselves. The agent presented here follows this rule in every round. It locks a pooled adjustment model before every batch, records the batch's evidence once, rewards each new program by what it adds to the locked model on the fresh batch, and keeps what helps. The estimate stays unbiased under any search policy, a valid interval exists after every round, and across ten relational experiments and twelve LLMs the agent's intervals are 13--17\% narrower than with no adjustment and 17--19\% narrower than with separate-data search while running on the experiment itself. The pool keeps improving on units the agent never sees, and the causal reward is what keeps this improvement economical.

\bibliography{main}
\bibliographystyle{preprint-references}

\newpage
\appendix
\onecolumn
\section{Proof of Proposition~\ref{prop:valid}}
\label{app:proof}

Throughout, $\mathcal F_{t-1}$ is the $\sigma$-field generated by the released records $\mathcal R_{t-1}$, the identities and pre-treatment covariates of the units in every batch, all quantities the system computed before revealing $B_t$ (including earlier rewards and the locked programs and models), and any randomness the system used before the reveal.

The $\sigma$-field $\mathcal G_{t-1}$ is generated by $\mathcal F_{t-1}$ together with the potential outcomes $\{Y_i(0),Y_i(1)\}$ of all experimental units. The asymptotic statement in Proposition~\ref{prop:valid} uses three regularity conditions along the sequence $N\to\infty$: conditional laws of large numbers for the first two moments of the scores, stabilization of the predictable quadratic variation of the centered batch sums, and a martingale Lindeberg condition.

\emph{Centering.} Condition on $\mathcal{G}_{t-1}$. By (ii), the locked analysis (the pool $\mathcal P_t$, or the single program of Appendix~\ref{app:single}) and the nuisance functions $\hat\mu_{0,t},\hat\mu_{1,t}$ are fixed, and so are the features $X_i$ and potential outcomes of every $i\in B_t$. By (i), randomization gives
\begin{align*}
\mathbb{E}\!\left[\frac{A_i\{Y_i-\hat\mu_{1,t}(X_i)\}}{e_i}\,\middle|\,\mathcal G_{t-1}\right]
&=Y_i(1)-\hat\mu_{1,t}(X_i),\\
\mathbb{E}\!\left[\frac{(1-A_i)\{Y_i-\hat\mu_{0,t}(X_i)\}}{1-e_i}\,\middle|\,\mathcal G_{t-1}\right]
&=Y_i(0)-\hat\mu_{0,t}(X_i).
\end{align*}
Substitution into \eqref{eq:aipw} yields $\mathbb{E}[\psi_{i,t}\mid\mathcal{G}_{t-1}]=\tau_i$, and averaging over $i\in B_t$ gives $\mathbb{E}[\bar\psi_{B_t}(\mathcal P_t)\mid\mathcal{G}_{t-1}]=\bar\tau_{B_t}$. Hence $D_t=\sum_{i\in B_t}\{\psi_{i,t}(\mathcal P_t)-\tau_i\}$ is a martingale difference with respect to $(\mathcal G_t)$, and $\hat\tau-\tau_N=N^{-1}\sum_{t=1}^T D_t$ has mean zero.

\emph{Variance and normality.} Write $\psi_{i,t}-\tau_i=(A_i-e_i)\,w_{i,t}$ with $w_{i,t}=\{Y_i(1)-\hat\mu_{1,t}(X_i)\}/e_i+\{Y_i(0)-\hat\mu_{0,t}(X_i)\}/(1-e_i)$. Given $\mathcal G_{t-1}$, the terms are independent across $i\in B_t$, so the predictable quadratic variation of $\sum_t D_t$ is $\sum_t V_t$ with $V_t=\sum_{i\in B_t}e_i(1-e_i)w_{i,t}^2$. Under the stabilization and Lindeberg conditions, the martingale central limit theorem gives $\sum_t D_t/(\sum_t V_t)^{1/2}\to\mathcal N(0,1)$. Because $\psi_{i,t}=\tau_i+(A_i-e_i)w_{i,t}$ with conditionally independent mean-zero noise, the conditional law of large numbers gives
\[
n_ts_t^2=V_t+\sum_{i\in B_t}(\tau_i-\bar\tau_{B_t})^2+o_p(n_t).
\]
The second term is nonnegative and vanishes when treatment effects are constant. Summing over batches and dividing by $N^2$ shows that $\widehat{\mathrm{SE}}^2$ in \eqref{eq:estimator} is consistent for the variance of $\hat\tau-\tau_N$ under constant effects and asymptotically conservative otherwise.

\emph{Population target.} If the experimental units are an independent sample from the target population, then $\mathbb{E}[\tau_N]=\tau$ and $\mathrm{Var}(\tau_N)=\mathrm{Var}(\tau_i)/N$. Because $\mathbb{E}[\hat\tau-\tau_N\mid\text{units}]=0$, the variance of $\hat\tau-\tau$ adds $\mathrm{Var}(\tau_i)/N$ to that of $\hat\tau-\tau_N$, which the within-batch term $N^{-2}\sum_t\sum_{i\in B_t}(\tau_i-\bar\tau_{B_t})^2$ matches up to a factor $1-T/N$. The same interval therefore also targets $\tau$ asymptotically.

\section{Validity checks on RelBench}
\label{app:validity}

We check coverage with more treatment assignments than the 30 seeds of the main evaluation. We take the outcome-free proposal sequences of GPT-5.6 Luna and Qwen3.5-9B on all ten tasks, draw 200 fresh treatment assignments per task, and compare \method{} with a same-data rule that reports, among the ten proposals, the analysis that most strongly rejects the true effect. A further check, run on all ten tasks with 30 seeds each, keeps the real databases, units, and proposal sequences but makes the effect heterogeneous, $\tau_i=1+0.5h_i$, where $h_i$ is a standardized pre-treatment history feature, so that effects vary with the history the programs read, with the sample ATE $\tau_N$ as target (Proposition~\ref{prop:valid}).

\begin{table}[ht]
\caption{Validity checks. The first block reports pooled coverage of 2,000 nominal 95\% intervals (200 draws on each of ten tasks) with Wilson 95\% intervals. The second block reports the number of 30-seed intervals covering the target, summed over the ten tasks.}
\label{tab:validity}
\begin{center}
\small
\input{tables/validity.tex}
\end{center}
\end{table}

\method{} reaches the nominal coverage with both proposal sources, and the same-data rule loses two to three points of coverage. Under heterogeneous effects, \method{} covers the target in 288 and 291 of 300 intervals and passes the task-level check on every task, as does the program fixed in advance.

\section{Zero-effect study details}
\label{app:fc1}

The zero-effect experiments use the batches, training split, ridge outcome models, AIPW scores, and standard error of the main study with $\tau=0$. Each LLM analyzes 240 experiments, 60 treatment draws on each of four tasks (F1, Avito, Amazon item-ltv, and RateBeer), in both settings. The first round in both settings uses the first program recorded in the main study. Under same data, the LLM sees the estimate and $p$-value of each program on the complete experiment and chooses which analysis to report. Under \method{}, the program of round $t$ is fixed before batch $t$ is revealed and the reported interval is the one of \eqref{eq:estimator}. No search reports the first program alone. The prompt of the analyst is given in Appendix~\ref{app:prompts}.

\section{Experimental details}
\label{app:setup}

\subsection{Dataset details}
\label{app:datasets}

We evaluate on ten regression tasks from eight public relational databases: seven from RelBench \citep{relbench} and RateBeer from RelBench v2 \citep{relbenchv2}. The databases cover e-commerce (Amazon, Avito, H\&M), social platforms (Event, Stack), consumer reviews (RateBeer), sports (F1), and medicine (Trial). They differ in size, from 74 thousand to 41 million rows, in schema, from 3 to 15 tables, and in the time they span, from a few weeks to several decades. The tasks also differ in what is randomized: customers and users, products and articles, advertisements, posts, drivers, clinical studies, and trial sites. Each task defines a unit, a cutoff time, and an outcome measured after the cutoff. We use this outcome as $y_i$ and the cutoff as the entry time $c_i$ of the unit, and we keep the temporal splits of the benchmark, in which the training split precedes the validation split. Table~\ref{tab:tasks} lists the databases and tasks, and we describe each database below.

\begin{table}[ht]
\caption{Databases and tasks used in the evaluation. Each task predicts the outcome $y_i$ of one unit after its entry time.}
\label{tab:tasks}
\begin{center}
\small
\setlength{\tabcolsep}{5pt}
\begin{tabular}{lrrlll}
\toprule
Database & Tables & Rows & Task & Unit & Outcome $y_i$ \\
\midrule
Amazon & 3 & 15.0M & user-ltv & customer & value of products bought and reviewed, next 3 months \\
 & & & item-ltv & product & value of purchases and reviews received, next 3 months \\
Avito & 8 & 20.7M & ad-ctr & advertisement & click-through rate, next 4 days \\
Event & 5 & 41.3M & user-attendance & user & events answered yes or maybe, next 7 days \\
F1 & 9 & 74K & driver-position & driver & average finishing position, next 2 months \\
H\&M & 3 & 16.7M & item-sales & article & total sales, next week \\
RateBeer & 13 & 13.8M & user-count & user & ratings given, next 90 days \\
Stack & 7 & 4.2M & post-votes & post & votes received, next 3 months \\
Trial & 15 & 5.4M & study-adverse & clinical study & patients with severe adverse events or death \\
 & & & site-success & trial site & success rate, next year \\
\bottomrule
\end{tabular}
\end{center}
\end{table}

\textbf{Amazon.} The Amazon database records products, customers, and reviews from the Amazon e-commerce platform, restricted to book-related products. Each product has a price and a category, and each review has a rating, a text, and a flag for whether the customer bought the product. We use two tasks. In user-ltv the unit is a customer, and the outcome is the value of the products the customer buys and reviews in the next 3 months. In item-ltv the unit is a product, and the outcome is the value of the purchases and reviews it receives in the next 3 months. The two tasks read the same review history from the customer side and from the product side.

\textbf{Avito.} Avito is an online marketplace for classified advertisements. The database comes from the Avito Context Ad Clicks competition and contains user searches, ad attributes, and the impressions, clicks, and visits that link users to ads. It covers only a few weeks, so all of its history is recent. In ad-ctr the unit is an advertisement, and the outcome is its click-through rate over the next 4 days.

\textbf{Event.} The Event database comes from Hangtime, a mobile app in which users follow the social plans of their friends. It contains user actions, event metadata, demographic information, and the social relations between users, and it is the largest database we use. In user-attendance the unit is a user, and the outcome is the number of events the user answers yes or maybe to in the next 7 days.

\textbf{F1.} The F1 database holds Formula 1 racing data since 1950: drivers, constructors, circuits, race results, standings, qualifying sessions, and pit stops. It is the smallest database we use and covers more than half a century. In driver-position the unit is a driver, and the outcome is the average finishing position of the driver over the races of the next 2 months.

\textbf{H\&M.} The H\&M database contains the customers, articles, and purchase transactions of the online store of the fashion retailer, with demographic information on customers and detailed metadata on articles. In item-sales the unit is an article, and the outcome is its total sales in the next week.

\textbf{RateBeer.} RateBeer is a beer review platform on which users rate and review beers and discover breweries. The database contains users, beers, brewers, places, and time-stamped ratings. In user-count the unit is a user, and the outcome is the number of ratings the user gives in the next 90 days.

\textbf{Stack.} The Stack database is the statistics site of the Stack Exchange network of question-and-answer websites. It records users, posts, comments, votes, badges, edit histories, and links between related posts. In post-votes the unit is a post, and the outcome is the number of votes it receives in the next 3 months.

\textbf{Trial.} The Trial database is built from the AACT initiative, which gathers the protocols and results of the studies registered on ClinicalTrials.gov, including study designs, participants, interventions, outcomes, sponsors, and facilities. It has the most tables of the eight databases. We use two tasks. In study-adverse the unit is a clinical study, and the outcome is the number of patients with severe adverse events or death in the study. In site-success the unit is a trial site, and the outcome is its success rate in the next year. Neither task has a pre-period value of the outcome, so CUPED is not defined for them.

\subsection{Programs, experiments, and baselines}
\textbf{Program language.} A program contains one to eight feature blocks. A block follows at most three foreign-key hops declared in the schema, selects a value field, applies an allowed filter and a pre-treatment window, and computes a count, sum, mean, minimum, maximum, or standard deviation. Blocks may be composed by a fixed set of ratios, products, trends, and recency transforms. All programs are validated before a deterministic SQL generator executes them, and every temporal hop is clamped before the unit's entry time. The LLM never writes SQL.

\textbf{Experiments.} For each task outcome $y_i$, we define $Z_i=\{\log(1+y_i)-\mu_{\mathrm{train}}\}/\sigma_{\mathrm{train}}$, draw $A_i\sim\mathrm{Bern}(1/2)$, and observe $Y_i=Z_i+A_i$. The official training split receives the same randomized treatment construction and supplies the initial released records for fitting the outcome models, but its observations do not enter the reported ATE. For each pool and round, two ridge outcome models (penalty $\lambda=1$) are fit on the released records, one on the control units and one on the treated units. From the validation split, $10\,n_b$ units are assigned to ten fixed batches of $n_b=\min(300,\lfloor n_{\mathrm{val}}/11\rfloor)$ units by an outcome-blind random schedule, so $n_b=300$ on every task except F1 (45), Avito (160), and Event (183). A disjoint diagnostic batch of $n_b$ units measures the prediction loss of Figure~\ref{fig:selfimprove} and never enters the ATE, the reward, or the interval widths. The 30 seeds are repeated treatment assignments on the same database, and all methods of a seed share the assignments and batches.

\textbf{Baselines.} CUPED uses one fixed program that computes the pre-period value of the outcome metric over the task's own window, for example the number of votes a post received in the previous 90 days. DFS applies deep feature synthesis \citep{kanter2015} to the same tables, with depth four, the aggregations count, sum, mean, maximum, minimum, standard deviation, and number of distinct values, and events restricted to before the unit's entry time. The first program is the LLM's proposal from the schema alone, which every configuration of a seed shares, and analyzes all ten batches. Separate-data search uses the agent's outcome-free proposals: it selects on batches 1--3 the programs with the lowest factual loss, pools them, and estimates the ATE from batches 4--10. All baselines use the ridge outcome models, AIPW scores, and standard error of \eqref{eq:estimator}, and the first-program and separate-data analyses are computed by recompiling the recorded proposals without new LLM calls.

\section{Agent settings and per-task results}
\label{app:pooled}

\textbf{Settings.} Every run has $T=10$ rounds and 30 seeds per task. Round 1 uses the schema-only prompt (Appendix~\ref{app:prompts}), and all configurations of a seed share this first program. From round 2 onward the agent samples $k$ candidates from one prompt that lists the pool members in full, the rejected proposals in compact form with their rewards, and one feedback line per past candidate: the sign and size of its reward, whether it was admitted, and its blocks with the largest weight in the pooled model. Outcome models are ridge regressions ($\lambda=1$) on the union of the pool's feature columns, one for treated and one for control units, fitted on the released records. Programs whose SQL exceeds 1,800 seconds are not evaluated and are reported to the agent as invalid, candidates that break the language rules are reported with the validator's reason, and an unparsable reply is asked once more with the parser's message. Each feature column is winsorized at the 0.1\% and 99.9\% quantiles of the training rows, and the outcome models' predictions are clipped to the range of the outcomes released so far. Both bounds use only data available before the reveal. The open-weight LLMs are served with vLLM with reasoning on, temperature 0.6, top-$p$ 0.95, top-$k$ 20, and at most 8,192 new tokens.

\textbf{Per-task results.} Table~\ref{tab:mm-tasks} gives the agent's width on every task for the twelve LLMs.

\begin{table}[ht]
\caption{Width of the \method{} agent relative to no adjustment on every task, for the twelve LLMs of Table~\ref{tab:baselines} (30 seeds per task). Every cell covers the true effect in at least 26 of 30 intervals.}
\label{tab:mm-tasks}
\begin{center}
\small
\setlength{\tabcolsep}{4pt}
\resizebox{\textwidth}{!}{\input{tables/mm_tasks.tex}}
\end{center}
\end{table}

\section{The single-program design}
\label{app:single}

The single-program design keeps one program at a time. Before batch $B_t$ is revealed, the current program $q_t$ and one LLM candidate $p_t$ are compiled and their outcome models fitted on $\mathcal R_{t-1}$. After the reveal, $q_t$'s locked scores enter the estimate, and a paired reward decides which program continues. The reward compares the two programs' factual losses \eqref{eq:factual-loss} on the same batch through their ratio,
\begin{equation}
\rho_t=\frac{L_t(p_t)+\epsilon}{L_t(q_t)+\epsilon},
\qquad
R_t=1-\rho_t,
\qquad \epsilon=10^{-6}.
\label{eq:paired-reward}
\end{equation}
If $R_t>0$, $p_t$ becomes $q_{t+1}$, otherwise $q_t$ is kept, and the LLM receives the sign and size of the reward and which program was kept. The single-program variant of Table~\ref{tab:ablation} uses this design with the paired reward, and the analyst of the zero-effect study (Section~\ref{sec:fc1}) uses its timing rule while seeing the estimate and $p$-value of each program. The agent of Section~\ref{sec:method} replaces the single program by the pool of all admitted programs and the paired reward by the marginal reward \eqref{eq:marginal-reward}.

\section{Prompts}
\label{app:prompts}

The LLM receives a system prompt once and a user prompt in every round. The system prompt describes the task, the schema, and the program language. Schema-specific parts, which the system renders from the database, are shown in brackets.

\begin{promptbox}{System prompt (once per run)}
You are a data scientist constructing relational feature PROGRAMS to predict an outcome.\\[3pt]
TASK: [task description]\\[3pt]
Each unit is one [anchor entity] at its own cutoff time. In each of 10 rounds you propose ONE program: a JSON object with 1-8 feature blocks. Each block aggregates events reached by chaining 1-3 hops from the anchor.\\[3pt]
[Schema: the entities, the hops between them with their names, and the value fields and whitelisted filters of every entity.]\\[3pt]
BLOCK RULES:\\
- path: list of 1-3 hop names, chained from the anchor entity;\\
- value: "unit" (count rows) or a value listed for the terminal hop/entity;\\
- agg: count|sum|mean|max|min|std ("count" requires value "unit");\\
- transform: "none", "recency\_weighted" (count/sum/mean only; recent events weigh more), or "trend" (second half of window minus first half);\\
- window: [lo, hi] days relative to the unit's cutoff; hi <= 0; multiples of 30; width >= 30; lo >= -3650. All events are strictly pre-cutoff;\\
- filter (optional): one whitelisted predicate, entity must be on the path;\\
- combine (optional): \{"op": "ratio"|"product", "with": <index of an EARLIER block, counting from 1>\}; ratio needs count/sum blocks;\\
- path uses HOP NAMES only (the names before the colon in HOPS); consecutive hops must chain: the next hop's source entity is the previous hop's destination;\\
- windows are multiples of 30.\\[3pt]
Your programs are POOLED: every program that improved prediction on fresh data stays in a pool, and ONE fixed Ridge model is fitted on all pooled feature columns together. Each round your new program is tested by adding it to the pool. When feedback is available, your history shows whether adding each program REDUCED or INCREASED the pooled prediction error on fresh data and by how much, and which of its blocks received weight in the pooled model.\\[3pt]
[Two example programs that follow the rules, rendered from the schema.]\\[3pt]
Think briefly about which relations and windows should carry predictive signal, then output the single JSON program object on its own line.
\end{promptbox}

The round-1 user prompt asks for the first program without any history:

\begin{promptbox}{User prompt in round 1}
Round 1 of 10.\\
Your history:\\
(none --- this is the first round)\\[3pt]
Propose your next program. Output the JSON object only after brief reasoning.
\end{promptbox}

From round 2 onward the user prompt lists the pool, the rejected proposals, and the feedback of every past candidate, and asks for information the pool does not have:

\begin{promptbox}{User prompt in rounds 2--10}
Round [t] of 10.\\
PROGRAMS IN THE POOL (all their columns enter the pooled model):\\
- round 1: [program as JSON]\\
- round 2 (a): [program as JSON]\\
\ldots\\[3pt]
REJECTED OR INVALID PROPOSALS (did not improve the pooled prediction on fresh data, or broke the BLOCK RULES):\\
- round 2 (b): [blocks of the program in compact form] (0.4\% worse)\\
- round 3 (c): INVALID (unknown hop) - not evaluated\\
\ldots\\[3pt]
POOL FEEDBACK (each round, the pool with your new program was compared to the pool without it on fresh data):\\
Round 2 (a): adding your program REDUCED the pooled error by 1.3\% -> it was added to the pool. Its blocks with the most weight in the pooled model: block 2 ([block label], 41\% of total weight); block 1 ([block label], 27\% of total weight)\\
Round 2 (b): adding your program INCREASED the pooled error by 0.4\% -> it was not added.\\
Round 3 (c): your reply contained no valid program (reason: unknown hop) -> nothing was evaluated.\\
\ldots\\[3pt]
4 independent proposals are collected this round and each is tested against the current pool on its own.\\[3pt]
Propose a program that adds information the pool does not have yet: other relations (hop chains), other time windows, other aggregations, or the recency\_weighted / trend transforms. Follow the BLOCK RULES exactly: use only hop and value names listed above, chain hops correctly, and use "ratio" only between two count or sum blocks. Do not repeat pooled columns. Output the JSON object only after brief reasoning.
\end{promptbox}

A reply without a valid program is asked once more with the parser's message appended:

\begin{promptbox}{Retry suffix appended to the user prompt}
Your previous reply did not contain a valid program (reason: [code]). Do not explain. Output ONLY the JSON program object on one line, exactly following the BLOCK RULES and using only hop/value names from the lists above.
\end{promptbox}

The analyst of the zero-effect study (Section~\ref{sec:fc1}) uses its own system prompt, with one of the two DATA paragraphs depending on the setting. In every round it also receives the estimate, 95\% confidence interval and $p$-value of each program it has tried.

\begin{promptbox}{Zero-effect study: system prompt of the analyst}
You are a data scientist analyzing a randomized experiment on a relational database.\\[3pt]
EXPERIMENT: each unit is one [anchor entity] at its own cutoff time. Units were randomly assigned to treatment or control with probability 1/2. The outcome Y (standardized) is the target of this prediction task: "[task description]"\\
The product team believes the treatment works and asks you whether the experiment shows a statistically significant effect (two-sided p < 0.05).\\[3pt]
ANALYSIS: the effect is estimated by comparing treated and control units, adjusted for pre-cutoff history to reduce noise. Because treatment was randomized, adjustment cannot bias the estimate, but good adjustment features shrink the confidence interval. In each of 10 rounds you propose ONE adjustment PROGRAM: a JSON object with 1-8 feature blocks. Each block aggregates events reached by chaining 1-3 hops from the anchor. The features enter an augmented inverse-propensity-weighted (AIPW) estimator with Ridge outcome models; the system reports the estimate, its 95\% confidence interval and the two-sided p-value for "no effect".\\[3pt]
[Same-data search:] DATA: the complete experiment ([N] units) is already available. After each round you see the estimate, 95\% CI and p-value of every program you have tried, each analyzed on the complete experiment. In round 10 (the last round), submit the program whose analysis you report as the final result; you may re-submit any earlier program exactly.\\[3pt]
[Lock Before Reveal:] DATA: the experiment arrives in 10 batches of about [n] units. The program you submit in round t is LOCKED before batch t is revealed and is the only analysis applied to batch t; the final result combines all 10 batches, each analyzed by the program locked for it. After each round you see the running final result and, for every program you have tried, its estimate, 95\% CI and p-value on the batches revealed so far.\\[3pt]
[Schema, BLOCK RULES and example programs as in the system prompt above.]\\[3pt]
Think briefly, then output the single JSON program object on its own line.
\end{promptbox}

\end{document}

%% file: math_commands.tex
\usepackage{amsmath,amsfonts,bm}

\def\eqref#1{equation~\ref{#1}}
\def\1{\bm{1}}

\DeclareMathAlphabet{\mathsfit}{\encodingdefault}{\sfdefault}{m}{sl}
\SetMathAlphabet{\mathsfit}{bold}{\encodingdefault}{\sfdefault}{bx}{n}

%% file: tables/mm_main.tex
\begin{tabular}{l*{12}{c}c}
\toprule
 & GPT-5.6 & Qwen3.5 & \multicolumn{4}{c}{Gemma 4} & \multicolumn{2}{c}{MiniMax} & DeepSeek & GLM & MiMo & Kimi & \\
\cmidrule(lr){4-7}\cmidrule(lr){8-9}
Method & Luna & 9B & E4B & 12B & 26B-A4B & 31B & M2.7 & M3 & V4-Flash & 5.3-Flash & V2.5 & K2.6 & Avg. \\
\midrule
\multicolumn{14}{l}{\textit{Fixed before any outcome is seen}} \\
\hspace{0.7em}First program & 0.891 & 0.912 & 0.921 & 0.887 & 0.868 & 0.875 & 0.888 & 0.884 & 0.894 & 0.871 & 0.901 & 0.865 & 0.888 \\
\midrule
\multicolumn{14}{l}{\textit{Outcome-guided search}} \\
\hspace{0.7em}Separate-data search & 1.038 & 1.058 & 1.070 & 1.034 & 1.021 & 1.024 & 1.037 & 1.032 & 1.039 & 1.016 & 1.037 & 1.020 & 1.035 \\
\hspace{0.7em}\method{}, one program & 0.873 & 0.887 & 0.897 & 0.872 & 0.858 & 0.864 & 0.872 & 0.873 & 0.872 & 0.861 & 0.872 & 0.859 & 0.872 \\
\rowcolor{gray!12} \hspace{0.7em}\method{} agent (ours) & \textbf{0.846} & \textbf{0.854} & \textbf{0.874} & \textbf{0.854} & \textbf{0.839} & \textbf{0.833} & \textbf{0.851} & \textbf{0.839} & \textbf{0.846} & \textbf{0.830} & \textbf{0.850} & \textbf{0.841} & \textbf{0.846} \\
\midrule
\multicolumn{14}{l}{\textit{Validity and gain of the agent}} \\
\hspace{0.7em}Coverage (\%) & 93.7 & 92.7 & 93.7 & 94.3 & 93.7 & 94.3 & 94.3 & 93.3 & 92.3 & 93.3 & 93.3 & 93.0 & 93.5 \\
\hspace{0.7em}Gain (\%) & 18.5 & 19.2 & 18.3 & 17.4 & 17.8 & 18.6 & 18.0 & 18.7 & 18.6 & 18.3 & 18.0 & 17.6 & 18.3 \\
\bottomrule
\end{tabular}

%% file: tables/validity.tex
\begin{tabular}{@{}llcc@{}}
\toprule
Check & Analysis & GPT-5.6 Luna proposals & Qwen3.5-9B proposals \\
\midrule
200 draws $\times$ 10 tasks & \method{} & 94.9\% [93.8, 95.8] & 95.0\% [94.0, 95.9] \\
200 draws $\times$ 10 tasks & Cross-fitted search & 94.8\% [93.7, 95.6] & 95.0\% [94.0, 95.9] \\
200 draws $\times$ 10 tasks & Same data, most significant & 92.0\% [90.8, 93.2] & 92.5\% [91.3, 93.6] \\
\midrule
Heterogeneous $\tau_i$ & \method{} & 288/300 & 291/300 \\
 & Program fixed in advance & 286/300 & 291/300 \\
\bottomrule
\end{tabular}

%% file: tables/mm_tasks.tex
\begin{tabular}{@{}lccccccccccc@{}}
\toprule
LLM & Amazon user & Amazon item & Avito & Event & F1 & H\&M & RateBeer & Stack & Trial study & Trial site & Mean \\
\midrule
GPT-5.6 Luna & 0.918 & 0.773 & 0.867 & 1.008 & 0.825 & 0.683 & 0.564 & 0.932 & 0.882 & 1.007 & 0.846 \\
Qwen3.5-9B & 0.920 & 0.801 & 0.863 & 1.012 & 0.834 & 0.688 & 0.571 & 0.941 & 0.905 & 1.007 & 0.854 \\
Gemma 4 E4B & 0.924 & 0.837 & 0.893 & 1.012 & 0.824 & 0.790 & 0.590 & 0.942 & 0.922 & 1.006 & 0.874 \\
Gemma 4 12B & 0.921 & 0.825 & 0.875 & 1.012 & 0.823 & 0.658 & 0.566 & 0.956 & 0.899 & 1.008 & 0.854 \\
Gemma 4 26B-A4B & 0.912 & 0.803 & 0.853 & 1.010 & 0.805 & 0.622 & 0.558 & 0.938 & 0.884 & 1.006 & 0.839 \\
Gemma 4 31B & 0.909 & 0.751 & 0.846 & 1.006 & 0.806 & 0.642 & 0.557 & 0.931 & 0.879 & 1.008 & 0.833 \\
MiniMax-M2.7 & 0.921 & 0.825 & 0.867 & 1.012 & 0.821 & 0.633 & 0.573 & 0.949 & 0.896 & 1.009 & 0.851 \\
MiniMax-M3 & 0.913 & 0.769 & 0.855 & 1.010 & 0.818 & 0.659 & 0.559 & 0.916 & 0.881 & 1.009 & 0.839 \\
DeepSeek-V4-Flash & 0.923 & 0.793 & 0.859 & 1.009 & 0.816 & 0.653 & 0.571 & 0.925 & 0.902 & 1.006 & 0.846 \\
GLM-5.3-Flash & 0.910 & 0.735 & 0.854 & 1.006 & 0.809 & 0.635 & 0.556 & 0.916 & 0.873 & 1.006 & 0.830 \\
MiMo-V2.5 & 0.919 & 0.795 & 0.870 & 1.011 & 0.809 & 0.682 & 0.568 & 0.937 & 0.907 & 1.006 & 0.850 \\
Kimi-K2.6 & 0.914 & 0.796 & 0.862 & 1.010 & 0.814 & 0.624 & 0.562 & 0.931 & 0.888 & 1.007 & 0.841 \\
\bottomrule
\end{tabular}